\documentclass[10pt,twocolumn,letterpaper]{article}

\usepackage{cvpr}              

\definecolor{cvprblue}{rgb}{0.21,0.49,0.74}
\usepackage[pagebackref,breaklinks,colorlinks,allcolors=cvprblue]{hyperref}

\usepackage[accsupp]{axessibility} 
\usepackage{amsmath}
\usepackage{graphicx}
\usepackage{float}
\usepackage{bbold}
\usepackage{subcaption}
\usepackage{multicol}
\usepackage{multirow}

\usepackage{algorithm} 
\usepackage{algpseudocode} 
\usepackage{amssymb}

\usepackage{colortbl}
\definecolor{Gray}{gray}{0.9}
\newcommand{\drule}{\specialrule{0.2pt}{1pt}{1pt}%
            \specialrule{0.2pt}{0pt}{\belowrulesep}%
            }

\newcommand\blfootnote[1]{%
  \begingroup
  \renewcommand\thefootnote{}\footnote{#1}%
  \addtocounter{footnote}{-1}%
  \endgroup
}

\def\paperID{4395} 
\def\confName{CVPR}
\def\confYear{2025}

\title{CustomKD: Customizing Large Vision Foundation for Edge Model Improvement via Knowledge Distillation}

\author{
Jungsoo Lee \; Debasmit Das \; Munawar Hayat \; \\ 
Sungha Choi$^\dagger$ \; Kyuwoong Hwang \; Fatih Porikli \; \vspace{0.2cm}\\
Qualcomm AI Research$^\ddagger$ \\
\texttt{\footnotesize\{jungsool, debadas, hayat, sunghac, kyuwoong, fporikli\}@qti.qualcomm.com} \; \\
}

\begin{document}
\maketitle

\blfootnote{$^\dagger$ Corresponding author. \hspace{0.1cm} $^\ddagger$ Qualcomm AI Research is an initiative of Qualcomm Technologies, Inc.}

\begin{abstract}
We propose a novel knowledge distillation approach, CustomKD, that effectively leverages large vision foundation models (LVFMs) to enhance the performance of edge models (\eg, MobileNetV3).
Despite recent advancements in LVFMs, such as DINOv2 and CLIP, their potential in knowledge distillation for enhancing edge models remains underexplored. 
While knowledge distillation is a promising approach for improving the performance of edge models, the discrepancy in model capacities and heterogeneous architectures between LVFMs and edge models poses a significant challenge. 
Our observation indicates that although utilizing larger backbones (\eg, ViT-S to ViT-L) in teacher models improves their downstream task performances, the knowledge distillation from the large teacher models fails to bring as much performance gain for student models as for teacher models due to the large model discrepancy.
Our \textit{simple yet effective} CustomKD customizes the well-generalized features inherent in LVFMs to a given student model in order to reduce model discrepancies.
Specifically, beyond providing well-generalized original knowledge from teachers, CustomKD aligns the features of teachers to those of students, making it easy for students to understand and overcome the large model discrepancy overall.
CustomKD significantly improves the performances of edge models in scenarios with unlabeled data such as unsupervised domain adaptation (\eg, OfficeHome and DomainNet) and semi-supervised learning (\eg, CIFAR-100 and ImageNet), achieving the new state-of-the-art performances.
\end{abstract}

\vspace{-0.35cm}
\section{Introduction}
\label{sec:intro}
\vspace{-0.1cm}

Recent efforts in computer vision have focused on building large vision foundation models (LVFMs), with a substantial number of parameters, generally trained using large-scale pretaining datasets.
Owing to their well-generalized representation, LVFMs are widely known to achieve state-of-the-art performances on diverse downstream tasks~\cite{dinov2,clip,openclip,sam,swinv2}.
However, while LVFMs have demonstrated impressive performance, their utilization in resource-constrained real-world applications is challenging due to their high computational costs and extensive parameters, which often impede their deployment~\cite{foundation_robotics, foundation_autonomous, foundation_review}. 
Consequently, edge models, known for their computational efficiency, may remain the preferred choice for real-world applications, especially when considering deployment on mobile devices. 
However, these edge models typically offer limited performance, so finding a solution to improve their capabilities while maintaining their low computation costs is necessary.

While training edge models with more labeled data samples to improve their performance on a certain downstream task would be one viable solution, collecting more labeled samples is labor-intensive and expensive. 
Without further data collection, one feasible solution to tackle such a challenge is performing knowledge distillation (KD)~\cite{cc,rkd,logits,decoupled_kd,show_attend_distill} with LVFMs serving as teachers for the edge models.
This approach can enhance the performance of edge models to be comparable to that of LVFMs while maintaining their low computational costs.
Specifically, by leveraging unlabeled data samples, we can extract meaningful information from LVFMs and transfer it to edge models without incurring expensive data annotation costs.

However, one major challenge of using LVFMs as teachers is the model discrepancy between LVFMs and edge models, leading to differences in representation spaces. 
In this KD setting, the model discrepancy arises from two primary factors. 
First, LVFMs are built with a vast number of parameters in order to understand the massive amount of knowledge from large-scale pretraining datasets while edge models are built on only a limited number of parameters.
Second, the heterogeneous architectures (\ie, architectural difference) between students (\eg, CNN based) and teachers (\eg, ViT based) may also cause the model discrepancy.
To be more specific, prevalent LVFMs employ ViT-based architectures~\cite{dinov2,openclip,swin,swinv2,florence} while CNN-based edge models are favored~\cite{CNN_cloud,CNN_iot,CNN_smart_homes} for their fast inference speed.
Our preliminary experiments demonstrate that the existing KD methods fail to effectively improve the performance of student model when the model discrepancy is substantial (Section~\ref{sec:motivation}).
Specifically, when utilizing larger backbones of teacher models (\eg, changing the backbone from ViT-S to ViT-L), we observe that existing KD methods fail to further improve the performances of edge models as much as the improved performances of teachers.

To address this issue, we propose a \textit{simple yet effective} feature alignment method called CustomKD, which alternates between two stages: 1) feature customization and 2) knowledge distillation.
In the feature customization stage, we customize the well-generalized features of LVFMs to a given student model by aligning them to the representation space of the student using the student's head classifier.
In the KD stage, we encourage the student model to imitate two different features: 1) the task-general feature extracted from the frozen teacher, and 2) the customized task-specific feature obtained during the feature customization stage.
Since the student model, including its head classifier, is updated during the KD stage, we alternate these two stages to progressively enhance the student model by continuously improving task-specific features. 
Importantly, we do not alter the architectures or inference processes of edge models, allowing us to significantly improve their performance without increasing inference speed.
Although feature alignment has been widely adopted and explored in various studies~\cite{deep_coral,cafa,student_customed_KD,its_all_head,fitnet}, we want to emphasize that finding how to perform feature alignment using LVFMs as teachers in KD is underexplored and our technical novelty lies in finding answer to such a challenge. 


The major contributions of this paper are:
\begin{itemize}
    \item[$\bullet$] Although using larger backbones improves the downstream task performances of LVFMs, our experiment demonstrates that existing KD methods fail to bring as much performance gain for edge models as for LVFMs.
    \item[$\bullet$] We propose CustomKD, a knowledge distillation method that enables to leverage LVFMs with large backbones as teachers by overcoming the large model discrepancy.  
    \item[$\bullet$] CustomKD consistently improves the performances of edge models on various tasks (\eg, unsupervised domain adaptation and semi-supervised learning), without architectural changes nor increased inference speed.
\end{itemize}

\vspace{-0.0cm}
\section{Related Work}
\label{sec:related_work}
\vspace{-0.0cm}

\subsection{Large Vision Foundation Models}
\vspace{-0.08cm}
The unprecedented breakthroughs of large language models in natural language processing have motivated computer vision studies to build large models for computer vision tasks.
Due to such efforts, large vision foundation models (LVFMs) have received attention for their outstanding performances in diverse downstream tasks~\cite{dinov2,clip,openclip,sam,swinv2,florence}, even with a simple linear probing. 
In this work, the definition of LVFMs includes the image encoders that were pretrained with images only (\eg, DINOv2~\cite{dinov2}) and those of vision-language models (\eg, CLIP~\cite{clip}). 
DINOv2~\cite{dinov2} improves the performance of DINO~\cite{dino} by leveraging a large-scale pretraining dataset curated using their proposed data processing pipeline, achieving state-of-the-art performances in diverse computer vision tasks. 
Similarly, CLIP~\cite{clip,openclip} shows robust generalization performances on diverse zero-shot classification tasks.
However, as aforementioned, utilizing such LVFMs in real-world applications is challenging since they are generally built on millions of parameters, showing slow inference speed~\cite{foundation_robotics, foundation_autonomous, foundation_review}.
For this reason, we require edge models capable of demonstrating comparable performance to LVFMs while maintaining high computational efficiency in real-world applications.

\begin{figure*}[t]
    \centering
    \includegraphics[width=0.85\textwidth]{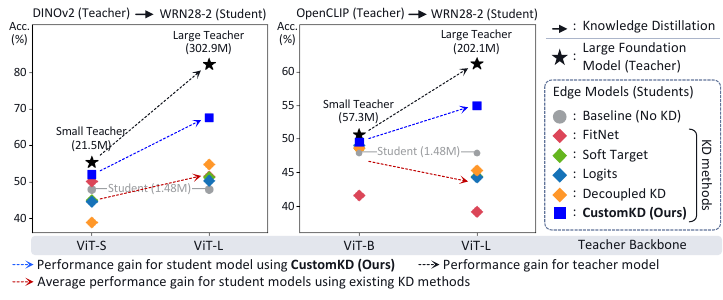}
    \vspace{-0.45cm}
    \caption{Limited performance gain with larger teachers. While utilizing small teachers (\eg, ViT-S, ViT-B) brings comparable or better performance than the teacher's performance, existing KD methods fail to further improve student's performance with large teachers (\textit{e.g.,} ViT-L). We use FitNet~\cite{fitnet}, Soft Target~\cite{soft_target}, Logits~\cite{logits}, and Decoupled KD~\cite{decoupled_kd} for conventional KD methods.}
    \label{fig:motivation}
    \vspace{-0.43cm}
\end{figure*}

\vspace{-0.12cm}
\subsection{Knowledge Distillation}
\vspace{-0.08cm}
Knowledge distillation (KD) is one viable solution to improve the performances of edge models (\ie, student models)~\cite{fitnet,cc,rkd,decoupled_kd,DearKD,does_kd_work,its_all_head,kd_from_strong_teacher,soft_target,logits,show_attend_distill,student_friendly_kd,student_friendly_teacher_kd,deit, tffd, diswot, norm}. 
KD is widely known for improving the performances of student models by teaching the hidden inter-class relationship at the prediction level~\cite{soft_target,logits,decoupled_kd} or improving the representation by imitating the features of teacher models~\cite{fitnet,cc,rkd}.
Due to this fact, we can utilize unlabeled datasets for further improving the performances of edge models using LVFMs as teachers~\cite{KD_LVFM_unlabeled}.
However, applying existing KD methods with LVFMs as teacher models brings limited performance gain due to the model discrepancy (Fig.~\ref{fig:motivation}), which was underexplored in previous work. 

Student-friendly knowledge distillation methods~\cite{on_the_efficacy_kd,takd, student_friendly_teacher_kd, its_all_head, student_customed_KD, one-for-all} have been proposed to address the discrepancy between teacher and student models.
For example, utilizing the frozen classifier of a pretrained student model for training teachers~\cite{its_all_head} could be one viable approach for the student-friendly knowledge distillation.
However, previous studies attempting to reduce the model discrepancy have limited their experiments to using similar architectures for the student and the teacher model (\eg, using ResNet architectures for both student and teacher).
While a recent KD study tries to overcome the heterogeneous architectures of students and teachers~\cite{one-for-all,wisdom_committee_kd}, they still only use small backbones of teacher models (\eg, ViT-S).
In this work, we first show that using large backbones (\eg, ViT-L) of LVFMs as teacher models is indeed challenging to further improve the performances of edge models with KD.  
Then, we propose CustomKD that enhances the performances of edge models even with large backbones of teacher models in heterogeneous architectures.

\vspace{-0.0cm}
\section{Method}
\label{sec:method}
\vspace{-0.05cm}
\subsection{Problem Setup}
\vspace{-0.1cm}
Throughout the paper, we define the student model as $\theta_s=(\theta^e_s, \theta^c_s)$ and the teacher model as $\theta_t=(\theta^e_t, \theta^c_t)$, where $\theta^e$ and $\theta^c$ represent the encoder and the head classifier, respectively.
Generally, knowledge distillation enforces the student model $\theta_s$ to imitate the teacher model $\theta_t$ either at the feature level~\cite{rkd,cc,fitnet} or at the prediction level~\cite{soft_target,logits,decoupled_kd}.
In this work, we assume that we have a pretrained $\theta_s$ that was trained with a small amount of labeled data $D_L$, composed of $\{x_i, y_i\}^{N_L}_{i=1} \in D_L$. 
Then, our main goal is to further improve the performance of $\theta_s$ by leveraging $\theta_t$ with an unlabeled data $D_U$, composed of $\{x_i, \}^{N_U}_{i=1} \in D_U$, additional to the labeled data $D_L$.
Under such a task setting, we mainly conduct experiments on unsupervised domain adaptation (\ie, UDA) and semi-supervised learning (\ie, SSL), where $D_U$ refers to the target dataset in UDA or the unlabeled dataset in SSL. 

\vspace{-0.15cm}
\subsection{Motivation}
\label{sec:motivation}
\vspace{-0.1cm}
Before diving into our proposed method, CustomKD, we first demonstrate the challenge of using LVFMs with large backbones (\eg, ViT-L) as teacher models through a preliminary experiment on the semi-supervised learning task using CIFAR-100 dataset with 400 labeled samples.
We use DINOv2~\cite{dinov2} and OpenCLIP~\cite{openclip} for the teacher models and WideResNet28-2~\cite{wrn} for the student model.

The gray-colored dots indicate the SSL performances of the edge model, and the black-colored stars indicate those of LVFMs trained with a simple linear probing. 
Our main observation is that the existing KD methods fail to further improve the performance at a comparable level to the teacher's performance when changing the backbone from a small one (\eg, ViT-S, ViT-B) to a larger one (\eg, ViT-L).
Specifically, the black dotted lines and the red dotted lines in Fig.~\ref{fig:motivation} indicate the performance gain of teacher models and the averaged performance gain of student models, respectively, by changing the backbone of teacher models.
As shown, the slopes of black lines are much steeper than those of red lines, indicating that the student models fail to improve their performances as much as teachers despite the large backbone of teachers used for KD.
We conjecture that the main reason for this observation is due to the significant increase in model discrepancy when changing the teacher with a larger backbone, hindering further performance gain.
Our proposed method, CustomKD, enables further performance gains even when using large backbones of LVFMs, as shown in blue squares and blue dotted lines in Fig.~\ref{fig:motivation}.

\begin{figure*}[t]
    \centering
    \includegraphics[width=0.85\textwidth]{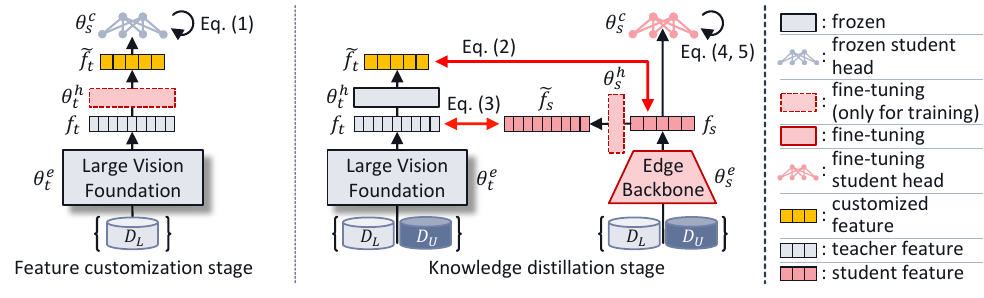}
    \vspace{-0.4cm}
    \caption{Overall framework of CustomKD. In the feature customization stage, we customize the well-generalized features of LVFMs to a given edge model using its head classifier ($\theta^c_s$). In the KD stage, we enforce the edge model to imitate the 1) task-general feature and 2) customized task-specific feature from the teachers. We alternate these two stages every epoch throughout the training process.}
    \label{fig:method_1}
    \vspace{-0.55cm}
\end{figure*}

\vspace{-0.1cm}
\subsection{Proposed Method}
\vspace{-0.1cm}
\noindent \textbf{Overall procedure}
Fig.~\ref{fig:method_1} depicts the overall process of CustomKD. 
We have two different stages that alternate throughout the training process: 1) feature customization stage and 2) knowledge distillation stage.
At a high level, during the feature customization stage, the well-generalized original features of the teacher model, $f_t$, are transformed into easily comprehensible features for the student model, $\tilde{f_t}$, by leveraging the classifier, $\theta^c_s$, shared from the student model.
Then, during the KD stage, we encourage $f_s$ to learn knowledge from both $f_t$ and $\tilde{f_t}$ by using MSE loss. 

\noindent \textbf{Feature customization stage}
The main purpose of this stage is to customize the well-generalized features of LVFMs to easily comprehensible features for a given student model.
We extract the original feature of the teacher model, $f_t=\theta^e_t(x)$, and forward to a projection layer $\theta^h_t$ in order to obtain $\tilde{f_t}=\theta^h_t(f_t)$.
During this process, assuming that we have a pretrained student model $\theta_s$, we replace the head classifier of the teacher model with the head classifier of the student model, denoted as $\theta^c_s$. 
We forward $\tilde{f_t}$ to the frozen $\theta^c_s$ and only update the projection layer $\theta^h_t$ using the labeled data, $(x, y) \in D_L$, formulated as,

\begin{equation}
    \min_{\theta^h_t} \hspace{0.1cm} \mathcal{L}_{t} = CE(\theta^{c}_{s}(\tilde{f_t}), y).
\vspace{-0.1cm}
\end{equation}
Through this process, we obtain $\tilde{f_t}$, the feature customized to a given student model derived from the well-generalized representation of the teacher model. 
Here, we do not use unlabeled data for this stage since our main goal is to use the well-generalized representation of $\theta^e_t$, rather than obtaining the optimal parameters of $\theta^h_t$ that adapted well to the unlabeled data (\eg, target domain for UDA task).

\noindent \textbf{Knowledge distillation stage}
During this stage, we encourage $f_s$ to learn knowledge from two different features: 1) $f_t$, which is the original feature from the teacher model, and 2) $\tilde{f_t}$, which is the feature we obtained during the feature customization stage. 
Intuitively, the two features contain different knowledge. 
While $f_t$ preserves the knowledge that the teacher model learned during the pretraining stage without loss of information, it only contains the task-general knowledge. 
On the other hand, while $\tilde{f_t}$ includes task-specific knowledge customized for promoting the understanding of a given student model, it inevitably has loss of information due to the projection layer $\theta^h_t$.
Our goal is to encourage the student model to learn both task-specific and task-general knowledge from the two different features, which the necessity of each knowledge is demonstrated in Table~\ref{tab:hyper_param}. 

In this stage, we use both labeled and unlabeled datasets, $x \in \{D_L, D_U\}$.
Since the customized feature $\tilde{f_t}=\theta^h_t(f_t)$ has the same embedding dimension with $f_s$, the supervision for imitating the task-specific feature $\tilde{f_t}$ is formulated as,
\begin{equation}
    \min_{\theta^e_s} \hspace{0.1cm} \mathcal{L}_{\tilde{f_t}} = ||f_s - \tilde{f_t}||^{2}.
\vspace{-0.1cm}
\end{equation}

Regarding the supervision on task-general feature $f_t$, $f_t$ generally have different embedding dimensions with $f_s$, so we forward $f_s$ to a projection layer $\theta^h_s$ and obtain $\tilde{f_s}=\theta^h_s(f_s)$ for imitating $f_t$, which is formulated as,
\begin{equation}
    \min_{\theta^h_s, \theta^e_s} \hspace{0.1cm} \mathcal{L}_{f_t} = ||\tilde{f_s} - f_t||^{2}.
\vspace{-0.1cm}
\end{equation}

\begin{table*}[t]
\centering
\begin{center}
{\resizebox{0.925\textwidth}{!}{
{
\begin{tabular}{c|c|cccccccccccc|c} 
\toprule
Category & Method & A2CA & A2P & A2RW & CA2A & CA2P & CA2RW & P2A & P2CA & P2RW & RW2A & RW2CA & RW2P & Avg. \\ 
\drule
 & DINOv2~\cite{dinov2}                       & 73.95 & 87.07 & 88.50 & 82.53 & 86.84 & 86.21 & 73.55 & 68.29 & 85.61 & 84.51 & 75.58 & 92.95 & 82.13 \\
& MobileNetV3~\cite{mobilenet}                & 37.57 & 46.74 & 58.39 & 33.79 & 48.50 & 48.73 & 32.92 & 35.12 & 56.69 & 49.03 & 42.36 & 66.84 & 46.39 \\

\midrule

\multirow{4}{*}{UDA} & DAN~\cite{dan}         & 33.68 & 43.34 & 54.62 & 31.81 & 46.86 & 48.22 & 31.69 & 35.21 & 55.27 & 47.01 & 40.76 & 64.16 & 44.39 \\
& Deep Coral~\cite{deep_coral}                & 34.16 & 44.04 & 54.49 & 31.60 & 46.79 & 48.18 & 31.69 & 35.19 & 55.36 & 46.64 & 40.64 & 64.02 & 44.40 \\
& DANN~\cite{dann}                            & 36.54 & 45.69 & 58.34 & 33.75 & 47.67 & 49.19 & 32.01 & 35.21 & 56.32 & 48.54 & 43.14 & 66.86 & 46.11 \\
& DSAN~\cite{dsan}                            & 37.73 & 47.08 & 58.94 & 34.98 & 48.93 & 50.75 & 33.75 & 37.27 & 58.73 & 50.14 & 44.08 & 67.90 & 47.52 \\
\midrule
& CC~\cite{cc}                                & 36.04 & 50.82 & 60.73 & 37.29 & 52.26 & 52.90 & 33.66 & 36.40 & 59.95 & 51.30 & 43.39 & 70.85 & 48.80 \\
Feature & RKD~\cite{rkd}                      & 44.51 & 58.30 & 63.99 & 40.34 & 59.52 & 57.22 & 37.37 & 46.46 & 64.59 & 56.94 & 54.00 & 76.41 & 54.97 \\
KD & FitNet~\cite{fitnet}                     & 55.05 & 69.36 & 72.92 & 55.42 & 68.19 & 70.12 & 50.35 & 55.92 & 76.36 & 68.52 & 63.51 & 83.98 & 65.81 \\
&  \cellcolor{Gray}CustomKD                      & \cellcolor{Gray}58.65 & \cellcolor{Gray}72.65 & \cellcolor{Gray}74.57 & \cellcolor{Gray}62.67 & \cellcolor{Gray}76.08 & \cellcolor{Gray}74.04 & \cellcolor{Gray}57.97 & \cellcolor{Gray}56.72 & \cellcolor{Gray}78.08 & \cellcolor{Gray}71.61 & \cellcolor{Gray}63.83 & \cellcolor{Gray}85.45 & \cellcolor{Gray}\textbf{69.36} \\

\cmidrule{1-15}

& Soft Target~\cite{soft_target}              & 53.33 & 63.08 & 69.57 & 50.60 & 64.9 & 62.52 & 44.75 & 49.67 & 67.11 & 63.33 & 59.61 & 80.47 & 60.75 \\
Prediction & Logits~\cite{logits}             & 58.47 & 73.06 & 76.68 & 64.24 & 75.74 & 73.15 & 56.7 & 58.79 & 73.93 & 73.38 & 64.15 & 85.90 & 69.52 \\
KD & DKD~\cite{decoupled_kd}                  & 60.27 & 72.47 & 76.91 & 63.25 & 75.74 & 72.32 & 55.62 & 57.89 & 74.71 & 73.96 & 64.63 & 85.94 & 69.48 \\
& \cellcolor{Gray}DKD~\cite{decoupled_kd} + CustomKD              & \cellcolor{Gray}63.60 & \cellcolor{Gray}77.95 & \cellcolor{Gray}80.17 & \cellcolor{Gray}70.70 & \cellcolor{Gray}80.63 & \cellcolor{Gray}79.00 & \cellcolor{Gray}64.36 & \cellcolor{Gray}64.40 & \cellcolor{Gray}80.67 & \cellcolor{Gray}77.59 & \cellcolor{Gray}67.90 & \cellcolor{Gray}88.60 & \cellcolor{Gray}\textbf{74.63} \\
\bottomrule
\end{tabular}}}}
\end{center}
\vspace{-0.6cm}
\caption{Image classification accuracy on OfficeHome. Bold digits indicate the best results in averaged accuracy among each category of knowledge distillation baselines.}
\vspace{-0.25cm}
\label{tab:main_uda_office} 
\end{table*}

\begin{table*}[t]
\centering
\begin{center}
{\resizebox{0.925\textwidth}{!}{
{
\begin{tabular}{c|c|cccccccccccc|c} 
\toprule
Category & Method & RW2CA & RW2P & RW2S & CA2RW & CA2P & CA2S & P2RW & P2CA & P2S & S2RW & S2CA & S2P & Avg. \\ 
\drule
 & DINOv2~\cite{dinov2}                        & 69.63 & 64.57 & 61.41 & 71.05 & 58.46 & 63.62 & 72.64 & 63.14 & 59.77 & 70.30 & 73.62 & 62.18 & 65.87 \\
& MobileNetV3~\cite{mobilenet}                & 38.43 & 35.83 & 24.54 & 39.76 & 24.36 & 28.59 & 45.29 & 33.42 & 26.40 & 38.93 & 43.16 & 30.46 & 34.10 \\

\midrule

\multirow{4}{*}{UDA} & DAN~\cite{dan}         & 32.47 & 32.83 & 21.54 & 36.03 & 21.59 & 24.28 & 40.84 & 26.90 & 22.61 & 34.31 & 35.28 & 27.12 & 29.65 \\
& Deep Coral~\cite{deep_coral}                & 32.55 & 32.86 & 21.39 & 36.00 & 21.57 & 24.23 & 40.87 & 26.90 & 22.46 & 34.34 & 35.20 & 27.07 & 29.62 \\
& DANN~\cite{dann}                            & 34.71 & 33.73 & 22.23 & 37.52 & 22.73 & 25.62 & 42.25 & 29.49 & 23.89 & 36.01 & 38.76 & 28.16 & 31.26 \\
& DSAN~\cite{dsan}                            & 38.74 & 37.20 & 26.36 & 39.88 & 25.25 & 29.94 & 44.57 & 32.76 & 27.37 & 38.40 & 42.74 & 30.73 & 34.50 \\
\midrule
& CC~\cite{cc}                                & 36.75 & 36.90 & 21.20 & 39.30 & 24.13 & 27.88 & 45.79 & 32.72 & 25.08 & 39.46 & 46.17 & 30.92 & 33.86 \\
Feature & RKD~\cite{rkd}                      & 38.63 & 36.91 & 24.31 & 37.28 & 23.94 & 29.57 & 44.15 & 33.58 & 26.70 & 36.41 & 45.80 & 31.00 & 34.02 \\
KD & FitNet~\cite{fitnet}                     & 43.08 & 42.42 & 27.99 & 41.71 & 26.24 & 34.59 & 48.83 & 37.78 & 29.14 & 42.04 & 50.91 & 34.11 & 38.24 \\
&  \cellcolor{Gray}CustomKD                       & \cellcolor{Gray}43.26 & \cellcolor{Gray}42.76 & \cellcolor{Gray}29.08 & \cellcolor{Gray}41.38 & \cellcolor{Gray}26.91 & \cellcolor{Gray}34.3 & \cellcolor{Gray}48.43 & \cellcolor{Gray}37.72 & \cellcolor{Gray}30.91 & \cellcolor{Gray}41.98 & \cellcolor{Gray}51.18 & \cellcolor{Gray}34.15 & \cellcolor{Gray}\textbf{38.51} \\

\cmidrule{1-15}

& Soft Target~\cite{soft_target}              & 42.88 & 39.07 & 28.45 & 40.81 & 26.46 & 31.66 & 46.06 & 36.75 & 29.39 & 39.45 & 47.67 & 32.24 & 36.74 \\
Prediction & Logits~\cite{logits}             & 39.36 & 36.69 & 27.16 & 36.90 & 23.08 & 30.56 & 42.72 & 30.52 & 26.49 & 32.49 & 46.05 & 26.97 & 33.25 \\
KD & DKD~\cite{decoupled_kd}                  & 45.00 & 41.40 & 31.70 & 39.91 & 26.82 & 34.58 & 46.53 & 36.61 & 31.26 & 38.85 & 50.92 & 32.34 & 37.99 \\
& \cellcolor{Gray}DKD~\cite{decoupled_kd} + CustomKD              & \cellcolor{Gray}47.01 & \cellcolor{Gray}43.44 & \cellcolor{Gray}35.88 & \cellcolor{Gray}45.15 & \cellcolor{Gray}31.25 & \cellcolor{Gray}36.42 & \cellcolor{Gray}48.91 & \cellcolor{Gray}38.63 & \cellcolor{Gray}33.79 & \cellcolor{Gray}43.90 & \cellcolor{Gray}52.82 & \cellcolor{Gray}36.00 & \cellcolor{Gray}\textbf{41.10} \\
\bottomrule
\end{tabular}}}}
\end{center}
\vspace{-0.6cm}
\caption{Image classification accuracy on DomainNet. Bold digits indicate the best results in averaged accuracy among each category of knowledge distillation baselines.}
\vspace{-0.55cm}
\label{tab:main_uda_domainnet} 
\end{table*}

Additionally, we use the standard cross entropy loss for the labeled data and entropy minimization for the unlabeled data, widely used in domain adaptation studies~\cite{woc, ecotta, lim2023ttn, swr, label_shift_adapter,shot}, in order to prevent the model from being overfitted to the labeled data.
The two loss functions are formulated as,
\begin{equation}
    \min_{\theta^c_s, \theta^e_s} \hspace{0.1cm} \mathcal{L}_{L} = CE(\theta_s(x_L), y_L),
\end{equation}
\begin{equation}
    \min_{\theta^c_s, \theta^e_s} \hspace{0.1cm} \mathcal{L}_{U} = H(\hat{\theta}_s(x_U)),
\end{equation}
where $H(p)=\Sigma^{C}_{k=1}p^{k}\log{p^{k}}$ with $C$ number of classes.
Then, our final loss function is formulated as follows,
\vspace{-0.1cm}
\begin{equation}
    \min_{\theta^h_s, \theta^c_s, \theta^e_s} \hspace{0.1cm} \mathcal{L}_{s} = \mathcal{L}_{L} + \lambda_{U}\mathcal{L}_{U} + \lambda_{f_t}\mathcal{L}_{f_t} + \lambda_{\tilde{f_t}}\mathcal{L}_{\tilde{f_t}}.
\end{equation}
\vspace{-0.1cm}
As aforementioned, CustomKD alternates the feature customization stage and the KD stage after each epoch.
In other words, we bring $\theta^c_s$ for the head classifier of the teacher during the feature customization stage after every epoch of the knowledge distillation stage.

Following are the advantages of CustomKD.
First, CustomKD is independent of the original training process of the edge model, so it improves the performance of any given off-the-shelf pretrained edge model.
Second, CustomKD does not incur additional computational costs during the inference stage as we discard $\theta^h_t$ and $\theta^h_s$, which are updated during the training phase.
Consequently, we believe that CustomKD can significantly improve the performance of a given off-the-shelf edge model without additional inference costs by leveraging LVFMs even with heterogeneous architectures distinct from the edge model.

\section{Experiments}
\label{sec:experiments}
\subsection{Experimental Setup}
\noindent \textbf{Evaluation settings}
As aforementioned, the main goal of this work is to utilize the unlabeled data for further improving downstream tasks of edge models leveraging LVFMs as the teacher models.
For this, we conduct experiments on tasks where unlabeled data are given: 1) unsupervised domain adaptation (UDA) and 2) semi-supervised learning (SSL). 
Then, we bring a pretrained off-the-shelf edge model and show that CustomKD consistently improves its performances across diverse tasks. 
Since our framework shows the effectiveness of utilizing large models as teachers in KD for tasks with unlabeled data, we compare CustomKD with baseline methods of both KD and each task.

\noindent \textbf{Datasets}
We use OfficeHome and DomainNet for the UDA task and CIFAR-100 and ImageNet for the SSL task.
For CIFAR-100, we vary the number of labeled samples to 400, 2500, and 10000, following the Unified SSL Benchmark (USB)~\cite{usb}. 
For ImageNet, we use subsets of 1\% and 10\% of labeled images and conduct comparisons with other existing baselines, following SimMatch~\cite{simmatch}. 

\noindent \textbf{Implementation details}
For the teacher models, we use DINOv2~\cite{dinov2} and OpenCLIP~\cite{openclip} for both tasks.
For the student models, we use  MobileNetV3~\cite{mobilenet} for the UDA task, and WideResNet28-2~\cite{wrn}, ResNet18~\cite{resnet}, and ResNet50~\cite{resnet} for the SSL task.
Regarding the KD baseline methods, we use the codes provided by the repository named Knowledge-Distillation-Zoo~\footnote{https://github.com/AberHu/Knowledge-Distillation-Zoo}.
For our experiments, we select conventional KD baseline methods that are categorized into prediction-level and feature-level methods.
For the prediction-level KD methods (\eg, Logits~\cite{logits}, Soft Targets~\cite{soft_target}, and Decoupled KD~\cite{decoupled_kd}), we perform linear probing and obtain a pretrained head classifier of teacher model, $\theta^c_t$.
For the feature-level KD methods (\eg, Relational KD~\cite{rkd}, Correlation Congruence~\cite{cc}, and FitNet~\cite{fitnet}), including CustomKD, we only use the frozen teacher model $\theta^e_t$, without training beforehand.
Beyond conventional KD methods, we further compare CustomKD with recent KD approaches, including TfFD~\cite{tffd}, NORM~\cite{norm}, and DisWot~\cite{diswot}, on OfficeHome to demonstrate the superiority of CustomKD.
For the UDA task, we follow the protocol of the repository named DeepDA~\footnote{https://github.com/jindongwang/transferlearning/tree/master/code/DeepDA}.
Regarding the SSL, we follow USB~\cite{usb} for CIFAR-100 and SimMatch~\cite{simmatch} for ImageNet.
Since we observe a performance gap between the reported values of USB and our obtained results by running the codes of USB on CIFAR-100, we additionally report our obtained results by denoting $*$ next to the results. 

\begin{table}[t]
\centering
\begin{center}
{\resizebox{0.44\textwidth}{!}{
{
\begin{tabular}{c|ccc|c} 
\toprule
Method & CA2A & CA2P & CA2RW & Avg. \\ 
\drule

Source Only             &   33.79	& 48.50	& 48.73	& 43.67 \\
\midrule
TfFD~\cite{tffd}        &   37.82	& 53.14	& 53.45	& 48.14 \\
\cellcolor{Gray}TfFD+ CustomKD  &   \cellcolor{Gray}48.87	& \cellcolor{Gray}63.60	& \cellcolor{Gray}61.40 & \cellcolor{Gray}\textbf{57.96} \\
\midrule
NORM~\cite{norm}        &   49.03	& 63.73	& 64.84	& 59.20 \\
\cellcolor{Gray}NORM+ CustomKD  &   \cellcolor{Gray}57.40 & \cellcolor{Gray}72.52	& \cellcolor{Gray}70.03	& \cellcolor{Gray}\textbf{66.65} \\
\midrule
DisWot~\cite{diswot}    &   71.61	& 80.99	& 78.75	& 77.12 \\
\cellcolor{Gray}DisWot+ CustomKD  &   \cellcolor{Gray}71.94 & \cellcolor{Gray}81.30	& \cellcolor{Gray}79.39	& \cellcolor{Gray}\textbf{77.54} \\
\bottomrule
\end{tabular}}}}
\end{center}
\vspace{-0.5cm}
\caption{Comparisons of CustomKD with recent knowledge distillation methods on UDA task using OfficeHome, using clipart as the source domain.}
\vspace{-0.5cm}
\label{tab:main_uda_recent} 
\end{table}

\vspace{-0.1cm}
\subsection{Image Classification}
\vspace{-0.1cm}
\noindent \textbf{Unsupervised domain adaptation} 
Table~\ref{tab:main_uda_office} and Table~\ref{tab:main_uda_domainnet} compare the UDA performances of both UDA and conventional KD baseline methods on OfficeHome and DomainNet, respectively.
We use MobileNetV3 for the student model and DINOv2 (ViT-L) for the teacher model. 
CustomKD achieves the best performances compared to existing both UDA and feature-level KD baseline methods.
When comparing with the source model, we improve the UDA performance of the source model by a large margin (\eg, an average of 22.97\% performance gain for OfficeHome). 
Since CustomKD applies KD at the feature level, it is also applicable with prediction-level KD (\eg, Logits~\cite{logits}, Soft Targets~\cite{soft_target}, DKD~\cite{decoupled_kd}), which we show its applicability by using CustomKD on DKD~\cite{decoupled_kd}.
As shown, applying CustomKD on DKD achieves the best performances compared to other baseline methods.
Since DomainNet is a relatively more challenging dataset than OfficeHome, we bring relatively small performance gain for DomainNet compared to OfficeHome.
Still, applying CustomKD on DKD achieves an average of 7\% performance gain compared to the source model, which is the best UDA performance among existing baseline methods. 

Table~\ref{tab:main_uda_recent} compares CustomKD with recent KD baseline methods, further demonstrating its superiority.
For the experiments, we apply CustomKD to various recent KD methods, including TfFD~\cite{tffd}, NORM~\cite{norm}, and DisWot~\cite{diswot}. 
As shown, CustomKD is applicable to an arbitrarily given KD method and shows consistent performance improvements. 
This result clearly highlights the effectiveness of CustomKD, even when compared with recent KD methods.


\noindent \textbf{Semi-supervised learning}
We also demonstrate the effectiveness of CustomKD on SSL, another task improving performances with unlabeled datasets.
Table~\ref{tab:main_semi_cifar100} compares the error rates of CIFAR-100 on both SSL and KD methods. 
We apply KD methods on WideResNet28-2~\cite{wrn} pretrained with AdaMatch~\cite{adamatch}.
While we report the results of AdaMatch that we obtain by running the codes of USB, we use the mean results of other SSL methods reported in USB.
We only report the recent state-of-the-art SSL methods while we show the results including other SSL methods in our Supplementary. 
As shown, we achieve the best error rates of CIFAR-100 on AdaMatch, outperforming the existing SSL methods. 
We want to emphasize that CustomKD does not require sophisticated techniques such as strong data augmentations or thresholding, which are required in various previous SSL studies~\cite{adamatch,simmatch,flexmatch,CoMatch,usb}.
Additionally, due to the page limit, we provide experimental results on ImageNet in our Supplementary to demonstrate the superiority of CustomKD over other SSL studies.

\section{Analysis}
\label{sec:analysis}
\subsection{Enhancing Knowledge Distillation Across Backbone Scales}
The preliminary experiment in Sec.~\ref{sec:motivation} motivates our work to propose CustomKD that overcomes the large model discrepancy between the student and the teacher model.
To demonstrate that CustomKD indeed improves the performance of the edge model even with the large backbone of teacher models, we conduct experiments with both small and large backbones of teacher models.
Under the SSL task, we use WideResNet28-2~\cite{wrn} and ResNet18~\cite{resnet} for the student models and DINOv2~\cite{dinov2} and OpenCLIP~\cite{openclip} for the teacher models.
We mainly compare with using $\mathcal{L}_{f_t}$ to demonstrate the necessity of training with $\mathcal{L}_{\tilde{f_t}}$ when the model discrepancy is substantial.

\begin{table}[!t]
  \centering
    \scalebox{0.9}{
    \begin{tabular}{c|c|ccc} 
    \toprule
    \multirow{2}{*}{Category} & \multirow{2}{*}{Methods} & \multicolumn{3}{c}{Labels} \\ 
     & & 400 \hfill & 2500 \hfill & 10000 \hfill \\ 
    \drule
    \multirow{6}{*}{SSL} & FixMatch~\cite{fixmatch}	        & 53.37 & 34.29 & 28.28 \\
    & SimMatch~\cite{simmatch}	        & 48.82 & 32.54 & 26.42 \\
    & FreeMatch~\cite{freematch}	    & 49.24 & 32.79 & 27.17 \\
    & SoftMatch~\cite{softmatch}	    & 49.64 & 33.05 & 27.26 \\
    & AdaMatch*~\cite{adamatch}	    & 52.07 & 37.92 & 32.5 \\
    \midrule
    \multirow{7}{*}{KD} & Soft Target~\cite{soft_target} & 48.71 & 31.73 & 27.66 \\
    & Logits~\cite{logits} & 49.71 & 33.42 & 28.16 \\
    & DKD~\cite{decoupled_kd} & 45.18 & 30.43 & 26.19 \\        
    & RKD~\cite{rkd} & 50.11 & 34.24 & 29.11 \\
    & CC~\cite{cc} & 49.85 & 33.72 & 28.75 \\
    & FitNet~\cite{fitnet} & 48.58 & 30.87 & 29.41 \\
    & \cellcolor{Gray}CustomKD & \cellcolor{Gray}\textbf{32.51} & \cellcolor{Gray}\textbf{25.52} & \cellcolor{Gray}\textbf{24.66} \\
    \bottomrule
    \end{tabular}}
  \vspace{-0.14cm}
  \caption{Error rates of semi-supervised learning on CIFAR-100. * indicates reproduced results using codes of USB benchmark.}
  \vspace{-0.45cm}
  \label{tab:main_semi_cifar100}
\end{table}

\begin{table*}[t]
\centering
\begin{center}
{\resizebox{0.8\textwidth}{!}{
{
\begin{tabular}{ccc|c|cc} 
\toprule
Teacher & Teacher & Teacher & \multirow{2}{*}{Methods} & WideResNet28-2~\cite{wrn} \space & ResNet18~\cite{resnet} \space \\
Type & Backbone & Error Rate* & & (1.48M, 0.22G) & (11.23M, 0.04G) \\ 
\drule
 - & - & - & Source & 52.07 & 73.10 \\

\midrule
\multirow{4}{*}{DINOv2~\cite{dino}} & ViT-S & \multirow{2}{*}{44.69} & $\mathcal{L}_{f_t}$ & 49.88 & 63.50 \\
& (21.52M, 5.52G) & & \cellcolor{Gray}$\mathcal{L}_{f_t} + \mathcal{L}_{\tilde{f_t}}$ & \cellcolor{Gray}\textbf{47.97 (-1.91)} & \cellcolor{Gray}\textbf{53.22 (-10.28)} \\
\cmidrule{2-6}
& ViT-L & \multirow{2}{*}{17.92} & $\mathcal{L}_{f_t}$ & 48.58 & 50.89 \\
& (302.91M, 77.82G) & & \cellcolor{Gray}$\mathcal{L}_{f_t} + \mathcal{L}_{\tilde{f_t}}$ & \cellcolor{Gray}\textbf{32.51 (-16.07)} & \cellcolor{Gray}\textbf{46.73 (-4.16)} \\

\midrule
\multirow{4}{*}{OpenCLIP~\cite{openclip}} & ViT-B & \multirow{2}{*}{49.38} & $\mathcal{L}_{f_t}$ & 58.46 & 75.51 \\
& (57.26M, 11.27G) & & \cellcolor{Gray}$\mathcal{L}_{f_t} + \mathcal{L}_{\tilde{f_t}}$ & \cellcolor{Gray}\textbf{50.49 (-7.97)} & \cellcolor{Gray}\textbf{58.56 (-16.95)} \\
\cmidrule{2-6}
& ViT-L & \multirow{2}{*}{38.78} & $\mathcal{L}_{f_t}$ & 60.89 & 75.70 \\
& (202.05M, 51.89G) & & \cellcolor{Gray}$\mathcal{L}_{f_t} + \mathcal{L}_{\tilde{f_t}}$ & \cellcolor{Gray}\textbf{45.02 (-15.87)} & \cellcolor{Gray}\textbf{56.55 (-19.15)} \\

\bottomrule
\end{tabular}}}}
\end{center}
\vspace{-0.55cm}
\caption{Error rates of CIFAR-100 using 400 labeled samples. Brackets indicate the number of parameters and Multiply-Accumulate Operations (MACs),.
* indicates that we performed linear probing using only labeled samples for the teacher.}
\vspace{-0.45cm}
\label{tab:main_ssl_diverse} 
\end{table*}

In Table~\ref{tab:main_ssl_diverse}, we observe that using $\mathcal{L}_{\tilde{f_t}}$ additional to $\mathcal{L}_{f_t}$ consistently improves the SSL performances of edge models, regardless of the backbones of the teacher models.
Specifically, the performance gap between the two loss functions generally enhances as the model discrepancy increases.
For example, with WideResNet28-2 as the student model, the performance gap increases from 1.91 to 16.07 for DINOv2 and from 7.97 to 15.87 for OpenCLIP by using larger backbones of teacher models.
The reason behind such an observation is as follows.
Using $\mathcal{L}_{f_t}$ simply projects the feature of the student to imitate that of the teacher. 
Simply utilizing a projection layer may limit fully understanding the knowledge of the teacher, especially when the model discrepancy is substantial.
On the other hand, using $\mathcal{L}_{\tilde{f_t}}$ additional to $\mathcal{L}_{f_t}$ not only teaches the task-general knowledge but also the task-specific knowledge customized for the student model, further boosting the performance of student regardless of the model discrepancy. 

\begin{figure}[b]
    \vspace{-0.3cm}
    \centering    
    \includegraphics[width=0.48\textwidth]{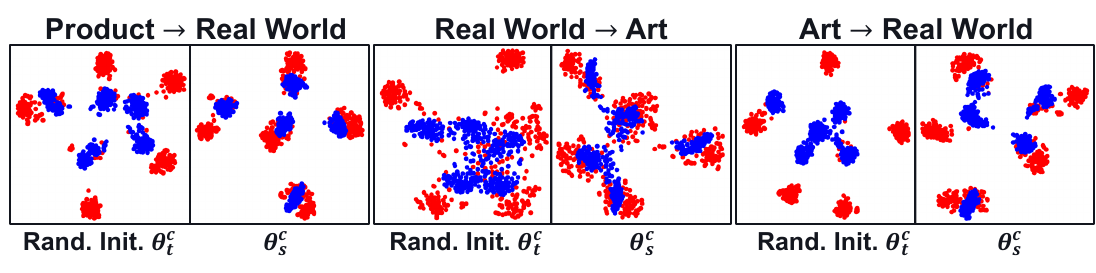}
    \vspace{-0.5cm}
    \caption{t-SNE visualization of $\tilde{f_t}$ (red) and $f_s$ (blue) on OfficeHome. For each domain, the left and right indicates training with randomly initialized $\theta^c_t$ and $\theta^c_s$, respectively, for the head classifier.}
    \vspace{-0.0cm}
    \label{fig:tsne}
\end{figure}

Additionally, student models may achieve better performance than the teacher model, even with a smaller number of parameters and lower computational costs.
The first and the second numbers in the bracket under each model indicate the number of parameters and Multiply-Accumulate Operations (MACs), respectively.
Applying CustomKD on WideResNet28-2 using DINOv2 with a large backbone (\ie, ViT-L) as the teacher achieves the error rate of 32.51\%. 
On the other hand, performing linear probing of DINOv2 with a small backbone (\ie, ViT-S) achieves the error rate of 44.69\%.
While the number of parameters and MACs of WideResNet28-2 are 1.48M and 0.22G, respectively, it achieves better performance compared to DINOv2 with ViT-S backbone, which has 21.52M number of parameters and 5.52G MACs.
Such a result clearly demonstrates the practicality of CustomKD when deploying such lightweight models on real-world applications with a comparable or better performance than teacher models.

\begin{table}[t]
\centering
\begin{center}
{\resizebox{0.47\textwidth}{!}{
{
\begin{tabular}{c|cccc} 
\toprule
Teacher & \multirow{2}{*}{Teacher Acc.} \space \space & \multirow{2}{*}{Student Acc.$\uparrow$} \space \space & \multirow{2}{*}{CKA($f_s$, $f_t$)$\uparrow$} \space \space & \multirow{2}{*}{CKA($f_s$, $\tilde{f_t}$)$\uparrow$} \\
Head Classifier & & & & \\
\drule
$\theta^c_t$ & 76.70 & 47.15 & 0.45 & 0.44 \\
\midrule
\cellcolor{Gray}$\theta^c_s$ & \cellcolor{Gray}75.48 & \cellcolor{Gray}\textbf{56.47} & \cellcolor{Gray}\textbf{0.54} & \cellcolor{Gray}\textbf{0.62} \\
\bottomrule
\end{tabular}}}}
\end{center}
\vspace{-0.45cm}
\caption{Comparisons on the initialization of the head classifier of teacher models during the feature customization stage. $\theta^c_t$ indicates randomly initialized head classifier.} 
\vspace{-0.45cm}
\label{tab:main_uda_random} 
\end{table}

\begin{table*}[!t]
    \begin{minipage}{.5\linewidth}
        \centering
        \begin{tabular}{ccc|c} 
        \toprule
        $\mathcal{L}_{L}$, $\mathcal{L}_{U}$ & \hspace{0.05cm} $\mathcal{L}_{f_t}$ & \hspace{0.02cm} $\mathcal{L}_{\tilde{f_t}}$ \hspace{0.025cm} & Avg. Acc. \\ 
        \drule
        \checkmark & \textendash & \textendash & 56.16 \\
        \checkmark & \checkmark & \textendash & 56.32 \\
        \checkmark & \textendash & \checkmark & 62.14 \\
        \cellcolor{Gray}\checkmark & \cellcolor{Gray}\checkmark & \cellcolor{Gray}\checkmark & \cellcolor{Gray}\textbf{62.76} \\
        \bottomrule
        \end{tabular}
        \vspace{0.0cm}
        \caption*{(a) KD loss functions.}        
        \label{tab:hyper_param_loss}
    \end{minipage}    
    \begin{minipage}{.5\linewidth}
        \centering
        \begin{tabular}{c|c} 
        \toprule
        Alternating Epochs & Avg. Acc. \\ 
        \drule
        30:1	& 62.22  \\  
        10:1	& 62.20  \\  
        5:1	    & 62.28  \\  
        \cellcolor{Gray}1:1	    & \cellcolor{Gray}\textbf{62.76}  \\  
        \bottomrule
        \end{tabular}
        \vspace{0.0cm}
        \caption*{(b) Alternating epochs for $\theta^h_t$}
        \label{tab:hyper_pram_epoch}
    \end{minipage}  
    \vspace{-0.3cm}
    \caption{Further studies on (a) our loss functions and (b) the frequency of the feature customization stage. We use MobileNetV3~\cite{mobilenet} and OpenCLIP~\cite{openclip} (ViT-B) for the student and the teacher model, respectively. We average the results of three target domains (\ie, art, clipart, and product) using the student model pretrained on the images of real world as the source domain in OfficeHome.}
    \vspace{-0.5cm}
    \label{tab:hyper_param}
\end{table*}

\subsection{Importance of Using Student Head Classifier}
During the feature customization stage, we bring $\theta^c_s$ after every epoch of knowledge distillation stage.
Table~\ref{tab:main_uda_random} compares such a design choice with utilizing a randomly initialized head classifier for the teacher model.
We conduct experiments on the UDA task on OfficeHome with MobileNetV3~\cite{mobilenet} and OpenCLIP~\cite{openclip} (ViT-L) for the student and the teacher, respectively.
In Table~\ref{tab:main_uda_random}, $\theta^c_t$ indicates the randomly initialized head classifier for the teacher model while $\theta^c_s$ refers to using the student head classifier for the teacher model. 
As shown, utilizing $\theta^c_t$ improves the UDA performance of the teacher model, denoted as Teacher Acc.
However, improving the performance of the teacher model does not guarantee a better performance of the student model, as supported  by previous studies~\cite{on_the_efficacy_kd,takd,student_customed_KD}. 
In other words, using $\theta^c_s$ for the head classifier of the teacher model improves the UDA performance of the student model, even with degraded performance of the teacher model. 

We further analyze such a result by measuring the centered kernel alignment (CKA)~\cite{cka} between 1) $f_s$ and $f_t$ and 2) $f_s$ and $\tilde{f_t}$.
CKA enables to compute the representation similarity of two matrices even with different embedding dimensions~\cite{vit_cnn,one-for-all}.
The higher the CKA values are, the similar the two matrices is. 
Table~\ref{tab:main_uda_random} shows that using $\theta^c_s$ for the head classifier of the teacher model increases both $\text{CKA}(f_s, f_t)$ and $\text{CKA}(f_s, \tilde{f_t})$. 
In other words, $\theta^c_s$ enables the student feature $f_s$ to imitate both the original task-general teacher feature $f_t$ and the customized task-specific teacher feature $\tilde{f_t}$ well.
Additionally, using $\theta^c_s$ further increases $\text{CKA}(f_s, \tilde{f_t})$ compared to $\text{CKA}(f_s, f_t)$, indicating that the representation of $\tilde{f_t}$ is aligned better with that of $f_s$ compared to $f_t$ due to the feature customization stage.

The t-SNE visualization~\cite{t-SNE} of $\tilde{f_t}$ and $f_s$ using 1) randomly initialized $\theta^c_t$ and 2) $\theta^c_s$ in Fig.~\ref{fig:tsne} also supports such a result.
The red and the blue dots indicate the projection of $\tilde{f_t}$ and $f_s$, respectively. 
While we use OfficeHome for the dataset of the visualization, we only select most frequent 5 categories among 65 categories to avoid information overload.
When utilizing the randomly initialized $\theta^c_t$, the blue dots misalign with the red dots, indicating the distinct representation spaces of $\tilde{f_t}$ and $f_s$.
On the other hand, when employing $\theta^c_s$, the blue dots align well with the red dots, indicating that the two representation spaces are similar and therefore demonstrating the necessity of using $\theta^c_s$ for the head classifier of the teacher model.

\begin{figure}[b]
    \vspace{-0.4cm}
    \centering    
    \includegraphics[width=0.4\textwidth]{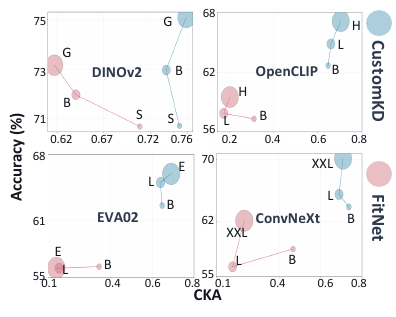}
    \vspace{-0.5cm}
    \caption{Consistent performance gains of using CustomKD compared to FitNet across diverse teachers and backbone scales.}
    \vspace{-0.0cm}
    \label{fig:cka}
\end{figure}

\vspace{-0.15cm}
\subsection{Ablation Studies}
\vspace{-0.1cm}
\noindent \textbf{Loss functions}
Table~\ref{tab:hyper_param}\textcolor{red}{(a)} shows the performance gain of using each loss function $\mathcal{L}_{f_t}$ and $\mathcal{L}_{\tilde{f_t}}$.
Additional to using only $\mathcal{L}_{L}$ and $\mathcal{L}_{U}$, individually utilizing $\mathcal{L}_{f_t}$ and $\mathcal{L}_{\tilde{f_t}}$ brings performance gain. 
When using the two KD loss functions together, we achieve the best performance.
While using $\mathcal{L}_{\tilde{f_t}}$ brings larger performance gain, this clearly demonstrates the necessity of learning both task-general knowledge from $\mathcal{L}_{f_t}$ and customized task-specific knowledge from $\mathcal{L}_{\tilde{f_t}}$.

\noindent \textbf{Alternating epochs}
Given that we perform the knowledge distillation stage every epoch, Table~\ref{tab:hyper_param}\textcolor{red}{(b)} investigates how the frequency of the feature customization stage influences the performance of the student model $\theta_s$.
For example, $5:1$ refers to updating $\theta^h_t$ in the feature customization stage after every 5 epochs of the knowledge distillation stage.
We observe that the performance degrades as $\theta^h_t$ is updated less frequently.
The main reason is as follows.
While $\theta_s$ is consistently updated during training, the teacher model may fail to provide adequate $\tilde{f_t}$ for $\theta_s$ if $\theta^h_t$ is less frequently updated, demonstrating the necessity of consistent update of $\tilde{f_t}$.
While the alternating epoch is a hyper-parameter, we emphasize that alternating the two stages itself is important for improving KD performances.

\vspace{-0.1cm}
\subsection{Consistent Performance Gains Across Teachers}
\vspace{-0.1cm}
Additionally, Fig.~\ref{fig:cka} demonstrates that CustomKD consistently improves KD performance across various large teacher backbones.
For our analysis, we conduct experiments using UDA with MobileNetV3 as the student model, OfficeHome (with Real World as the source domain) as the dataset, and FitNet~\cite{fitnet} for the baseline.
The x-axis and y-axis indicate the centered kernel alignment (CKA)~\cite{cka} and accuracy, respectively.
For computing CKA values, we use $f_s$ for the student model, while utilizing $f_t$ and $\tilde{f_t}$ for FitNet and CustomKD, respectively. 
As illustrated, CustomKD outperforms FitNet across diverse teacher models, including DINOv2~\cite{dinov2}, OpenCLIP~\cite{openclip}, EVA02~\cite{eva02}, and ConvNeXt~\cite{convnext}, spanning various backbone scales.

\vspace{-0.15cm}
\section{Future Work and Conclusions}
\label{sec:conclusion}
\vspace{-0.1cm}
In this work, we introduce CustomKD, a method that customizes the well-generalized features of large vision foundation models for a given edge model, aiming to further improve performances of edge models on downstream tasks.
Our preliminary experiment shows that existing KD methods bring limited performances gains of edge models, even when employing large backbones in teacher models, due to large model discrepancies. 
To address this issue, we propose aligning the representation of the teacher model to that of the student model by bringing the head classifier of the student. 
CustomKD alternates two stages: 1) feature customization stage that aligns the representation space of the teacher to that of the student and 2) knowledge distillation stage that encourages the feature of the student to imitate features of the teacher.  
Our work achieves new state-of-the-art performances on tasks with unlabeled data given, including UDA and SSL.
While our work mainly focuses on image classification tasks, we believe that this framework could also be applied to dense prediction tasks (\eg, semantic segmentation). 
To extend to such tasks, performing KD of spatial-wise knowledge should also be considered, which we leave for future work.
We believe that our work inspires future researchers to further endeavor to build fast and well-performing edge models by leveraging the rapidly developing large vision foundation models.

\newpage

\noindent \textbf{Acknowledgement}
We would like to thank Sungrack Yun, Simyung Chang, Hyunsin Park, Janghoon Cho, Juntae Lee, Hyoungwoo Park, Seokeon Choi, Kyuhong Shim, Seunghan Yang, Jihwan Bang, and Sunghyun Park of the Qualcomm AI Research team for their valuable discussions.

{
    \small
    \bibliographystyle{ieeenat_fullname}
    \bibliography{main}
}


\clearpage

\appendix

\section{Further Details of Experimental Setup}
\noindent \textbf{Datasets}
For the datasets in unsupervised domain adaptation (UDA), we use OfficeHome and DomainNet. 
OfficeHome contains four domains (images of art, clipart, product, real world) with 65 categories and 16,107 of images in total.
While DomainNet originally includes six domains, we conduct experiments on four domains, images of clipart, real world, sketch, and painting, following the previous studies~\cite{label_shift_adapter, sentry,tan2020classimbalanced}, with 345 classes and 373,061 of images in total. 
For the ImageNet in semi-supervised learning (SSL), we use 13000 and 128000 images during training for the evaluation settings of 1\% and 10\% of labeled samples, respectively, following SimMatch~\cite{simmatch}.

\noindent \textbf{Implementation details}
Eq.~5 in the main paper indicates the entropy minimization loss, $\mathcal{L}_{U}$, which we apply to all feature-level KD baseline methods~\cite{rkd,cc,fitnet} (including ours) using a consistent $\lambda_{U}$ value of 0.1 across all methods.
Since prediction-level KD baseline methods~\cite{logits,soft_target,decoupled_kd} have target predictions for $x_U$ obtained from the teacher models, we replace the entropy minimization loss with KD loss functions using those target predictions for them.

While using the default values of hyper-parameters of each baseline method, we find the lambda value for each KD loss function through a grid search of \{0.1, 1, 10, 100\}.
For UDA, we use the lambda value of 0.1 for Soft Targets~\cite{soft_target}, 1.0 for Logits~\cite{logits}, 0.1 for DKD~\cite{decoupled_kd}, 10.0 for RKD~\cite{rkd}, 10.0 for CC~\cite{cc}, and 10.0 for FitNet~\cite{fitnet}.
For CustomKD, while fixing 10.0 for $\lambda_{f_t}$, we set $\lambda_{\tilde{f_t}}$ as 10.0 and 1.0 for OfficeHome and DomainNet, respectively. 
For SSL, we use the lambda value of 0.1 for Soft Targets~\cite{soft_target}, 0.1 for Logits~\cite{logits}, 0.1 for DKD~\cite{decoupled_kd}, 0.1 for RKD~\cite{rkd}, 1.0 for CC~\cite{cc}, and 100.0 for FitNet~\cite{fitnet}.
For CustomKD, we set $\lambda_{f_t}=\lambda_{\tilde{f_t}}=100.0$ for CIFAR-100 and $\lambda_{f_t}=\lambda_{\tilde{f_t}}=10.0$ for ImageNet. 
For training KD baselines in UDA, we use the learning rate of 0.1 for both datasets, and we use 60 epochs and 20 epochs for OfficeHome and DomainNet, respectively. 
For SSL, we use the learning rate of 0.0001 for CIFAR-100 and 0.00005 for ImageNet with 50 epochs of training for both datasets. 
For linear probing, we train $\theta^c_t$ for 20 epochs for OfficeHome in UDA while using 10 epochs for DomainNet and CIFAR-100.
All experiments were conducted with a single GPU of RTX A5000 using less than 24GB GPU memory usage.

We fix the resolution size of the input image for teacher models to 224$\times$224 while using the image size of edge models by following the protocol we used.
This leads us to use 224$\times$224 for UDA and ImageNet in SSL, while using 32$\times$32 for CIFAR-100 in SSL.
Regarding the computations for the number of parameters and Multiply-Accumulate Operations (MACs) in Table~5 of the main paper, we use the repository named pytorch-OpCounter~\footnote{https://github.com/Lyken17/pytorch-OpCounter}.
Again, we compute the number of parameters and MACs by using the input size of 224$\times$224 and 32$\times$32 for the teacher and the student, respectively. 

Regarding $\theta^h_t$, we use a non-linear layer that consists of 1) a linear layer that transforms the embedding dimension of $f_t$ to that of $f_s$, 2) a batch normalization layer, and 3) ReLU function. 
For $\theta^h_s$, the input dimension and the output dimension for the first linear layer correspond to the embedding dimension of $f_s$ and $f_t$, respectively, while also using the batch normalization layer and ReLU function. 
We remove the batch normalization layers in both $\theta^h_t$ and $\theta^h_s$ for ImageNet in SSL.

\section{Pseudocodes for CustomKD}
Along with the description in Sec.~3.3 of the main paper, we also explain our proposed method CustomKD through pseudocodes. 
Stop gradient denotes that we do not allow gradient backpropagation for the process in the corresponding line. 
As mentioned in the main paper, we do not modify the forward process of the edge models during the inference stage. 

\begin{algorithm}
    \caption{Pseudocodes for CustomKD} 
    \begin{algorithmic}[1]
        \State \textbf{Require}: Pre-trained edge model $\theta_s$, pre-trained LVFM $\theta_t$, labeled dataset $D_L$, unlabeled dataset $D_U$
        
        \For {$epoch=1,2,\ldots,N$}
            \For {all $(x,y) \in D_L$}
                \Comment{Feature customization}
                \State Obtain $f_t$ = $\theta^e_t(x)$
                \Comment{Stop gradient}
                \State Obtain $\tilde{f_t}$ = $\theta^h_t(f_t)$
                \State Compute $\mathcal{L}_t=CE(\theta^c_s(\tilde{f_t}), y)$
                \State Update only $\theta^h_t$ via $\nabla\mathcal{L}_t $  
            \EndFor
            \For {all $(x,y) \in (D_L \cup D_U)$}
                \Comment{KD}
                \State Obtain $f_t$ = $\theta^e_t(x)$, $\tilde{f_t}$ = $\theta^h_t(f_t)$
                \Comment{Stop gradient}
                \State Obtain $f_s$ = $\theta^e_s(x)$, $\tilde{f_s}$ = $\theta^h_s(f_s)$

                \State Compute $\mathcal{L}_{\tilde{f_t}} = ||f_s - \tilde{f_t}||^{2}$,  $\mathcal{L}_{f_t} = ||\tilde{f_s} - f_t||^{2}$
                \State Compute $\mathcal{L}_{L} = CE(\theta_s(x_L), y_L)$, $\mathcal{L}_{U} = H(\hat{\theta}_s(x_U))$

                \State Compute $\mathcal{L}_s=\mathcal{L}_{L} + \lambda_{U}\mathcal{L}_{U} + \lambda_{f_t}\mathcal{L}_{f_t} + \lambda_{\tilde{f_t}}\mathcal{L}_{\tilde{f_t}}$
                \State Update $\theta^h_s$, $\theta^c_s$, $\theta^e_s$ via $\nabla\mathcal{L}_s $  
            \EndFor
        \EndFor
    \end{algorithmic} 
\end{algorithm}

\section{Further Experiments and Results}

In the main paper, we could not include the experiments of SSL task using ImageNet due to the page limit.
Table~\ref{tab:supple_semi_imagenet} shows that CustomKD again improves the SSL performance on ImageNet, demonstrating the scalability of CustomKD on large-scale datasets.
Again, we want to emphasize again that CustomKD does not require complicated techniques such as artificial label augmentation or prediction augmentations, which are widely used in SSL studies~\cite{adamatch,simmatch,flexmatch,CoMatch,usb}.
Therefore, we believe that the superiority of CustomKD is well established considering its consistent performance gains on various datasets and the simplicity of its training process.


In Table~\ref{tab:supple_semi_total}, we include the results of other SSL baseline methods on CIFAR-100 that we could not report due to the space limit in Table~4 of our main paper. 
We also compare our method with other KD baseline methods by using the pretrained model trained without any SSL module, denoted as Source.
Again, we observe that our method consistently improves the SSL performance on both edge models of AdaMatch and Source, demonstrating the wide scalability of our method regardless of the pretrained edge models.
Additionally, applying our method on Source model achieves a comparable performance with our reproduced AdaMatch.

Table~\ref{tab:supple_uda_diverse} shows that CustomKD also improves the UDA performances of other student models regardless of the backbone scale, following Table~5 of the main paper that reports the results on SSL. 
For the experiments, we use MobileNetV3~\cite{mobilenet} and ShuffleNet~\cite{shufflenet} for the edge models and OpenCLIP~\cite{openclip} for the teacher model. 
As shown, additionally using $\mathcal{L}_{\tilde{f_t}}$ consistently improves the UDA performance compared to using only $\mathcal{L}_{f_t}$.
Also, the performance gap between our method and using only $\mathcal{L}_{f_t}$ increases as we change the backbone of the teacher model from a small one (\ie, ViT-B) to a big one (\ie, ViT-L). 
Such a result demonstrates that it is challenging to improve the UDA performance with large backbones, while CustomKD can consistently improve it through customizing the well-generalized knowledge of teachers to a given edge model.

Due to the space limit in the main paper, only the average results are reported in Section~5. 
The results in Table~\ref{tab:supple_uda_random}, Table~\ref{tab:supple_uda_ablation_loss}, Table~\ref{tab:supple_uda_ablation_epoch} include the results of each domain in Table~6, Table~7(a), Table~7(b) of the main paper, respectively.

\begin{table}[!t]
  \centering
    \scalebox{0.99}{
    \begin{tabular}{c|cc|cc} 
    \toprule
    \multirow{2}{*}{Methods} & \multicolumn{2}{c|}{Top-1} & \multicolumn{2}{c}{Top-5} \\ 
     & 1\% & 10\% & 1\% & 10\% \\ 
    \drule
    Vat + EntMin~\cite{vat,minentp}	        &  - & 68.8 & -	& 88.5 \\
    S4L-Rotation~\cite{s4l}                 & -	 & 53.4 & -	& 83.8 \\ 
    UDA~\cite{UDA}                          & -	 & 68.8 & -	& 88.5 \\
    FixMatch~\cite{fixmatch}	            & -    & 71.5 & -	& 89.1 \\
    CoMatch~\cite{CoMatch}	                & 67.1 & 73.7 & 87.1 & 91.4 \\
    SimMatch~\cite{simmatch}	            & 67.2 &  74.4 & 87.1	& 91.6 \\
    \cellcolor{Gray}CustomKD        & \cellcolor{Gray}\textbf{67.5} & 	\cellcolor{Gray}\textbf{74.7} & \cellcolor{Gray}\textbf{87.6} & \cellcolor{Gray}\textbf{91.9} \\        
    \bottomrule
    \end{tabular}}
  \caption{Image classification accuracy of semi-supervised learning on Imagenet.}
  \vspace{-0.5cm}
  \label{tab:supple_semi_imagenet}
\end{table}

\section{Limitations}
One limitation, as previously mentioned in the conclusion of the main paper, is that this paper mainly focuses on addressing the image classification task.
While computer vision encompasses a plethora of tasks such as object detection or semantic segmentation, we leave them as the future work of our study.
However, we believe that our approach can be modified in a task-specific manner and applied to these tasks, potentially inspiring future researchers.

\begin{table*}[!t]
  \centering
  \vspace{-0.5cm}
    \scalebox{1.0}{
    \begin{tabular}{c|c|ccc} 
    \toprule
    \multirow{2}{*}{Category} & \multirow{2}{*}{Methods} & \multicolumn{3}{c}{Labels} \\ 
     & & 400 \hfill & 2500 \hfill & 10000 \hfill \\ 
    \drule
     & DINO-V2~\cite{dinov2}	                & 17.92 & 11.49 & 9.13 \\
     & Supervised*~\cite{wrn}	            & 28.20 & 28.20 & 28.20 \\
     \midrule
    \multirow{14}{*}{SSL} & Pseudolabel~\cite{pseudo-label}	    & 87.15 & 59.09 & 38.86 \\
    & Meanteacher~\cite{meanteacher}	    & 90.34 & 61.13 & 39.05 \\
    & Vat~\cite{vat}	                & 83.11 & 53.17 & 36.58 \\
    & MixMatch~\cite{mixmatch}	    & 79.95 & 49.58 & 32.10 \\
    & RemixMatch~\cite{remixmatch}	& 57.10 & 34.77 & 26.18 \\
    & AdaMatch~\cite{adamatch}	    & 47.82 & 33.26 & 27.53 \\
    & FixMatch~\cite{fixmatch}	    & 53.37 & 34.29 & 28.28 \\
    & FlexMatch~\cite{flexmatch}	    & 50.15 & 33.35 & 27.12 \\
    & Dash~\cite{dash}	                        & 53.98 & 34.47 & 27.72 \\
    & Crmatch~\cite{crmatch}	                        & 49.39 & 31.35 & 26.24 \\
    & CoMatch~\cite{CoMatch}	        & 60.98 & 37.24 & 28.15 \\
    & SimMatch~\cite{simmatch}	    & 48.82 & 32.54 & 26.42 \\
    & FreeMatch~\cite{freematch}	    & 49.24 & 32.79 & 27.17 \\
    & SoftMatch~\cite{softmatch}	    & 49.64 & 33.05 & 27.26 \\
    \midrule
    Pretrained & AdaMatch*~\cite{adamatch}	    & 52.07 & 37.92 & 32.5 \\
    \midrule
    \multirow{7}{*}{KD} & Soft Target~\cite{soft_target} & 48.71 & 31.73 & 27.66 \\
    & Logits~\cite{logits} & 49.71 & 33.42 & 28.16 \\
    & DKD~\cite{decoupled_kd} & 45.18 & 30.43 & 26.19 \\        
    & RKD~\cite{rkd} & 50.11 & 34.24 & 29.11 \\
    & CC~\cite{cc} & 49.85 & 33.72 & 28.75 \\
    & FitNet~\cite{fitnet} & 48.58 & 30.87 & 29.41 \\
    & \cellcolor{Gray}Ours & \cellcolor{Gray}\textbf{32.51} & \cellcolor{Gray}\textbf{25.52} & \cellcolor{Gray}\textbf{24.66} \\
    \midrule
    Pretrained & Source*~\cite{adamatch}	    & 90.36 & 69.73 & 49.60 \\
    \midrule
    \multirow{7}{*}{KD} & Soft Target~\cite{soft_target} & 66.67 & 61.77 & 41.81 \\
    & Logits~\cite{logits}      & 75.03 & 62.75 & 42.43 \\
    & DKD~\cite{decoupled_kd}   & 54.51 & 57.19 & 39.19 \\        
    & RKD~\cite{rkd}            & 92.20 & 64.00 & 42.73 \\
    & CC~\cite{cc}              & 91.90 & 63.62 & 41.76 \\
    & FitNet~\cite{fitnet}      & 68.00 & 49.40 & 37.72 \\    
    & \cellcolor{Gray}Ours & \cellcolor{Gray}\textbf{52.02} & \cellcolor{Gray}\textbf{38.59} & \cellcolor{Gray}\textbf{31.06} \\
    \bottomrule
    \end{tabular}}
  \caption{Error rates of semi-supervised learning on CIFAR-100. For the results of SSL methods, we report the mean results in Unified SSL Benchmark (USB)~\cite{usb}. * indicates reproduced results using codes of USB. We compare our method with other KD baselines using edge models pretrained 1) with AdaMatch and 2) without any SSL module, denoted as Source. Supervised indicates using all labels for the entire data samples.}
  \label{tab:supple_semi_total}
\end{table*}

\clearpage

\begin{table*}[t]
\centering
\begin{center}
{\resizebox{1.0\textwidth}{!}{
{
\begin{tabular}{ccc|c|cc} 
\toprule
Teacher & Teacher & Teacher & \multirow{2}{*}{Methods} & MobileNetV3~\cite{mobilenet} \space & ShuffleNet~\cite{shufflenet} \space \\
Type & Backbone & Accuracy* & & (2.54M, 0.06G) & (7.39M, 0.60G) \\ 
\drule
 - & - & - & Source & 52.70 & 62.16 \\

\midrule
\multirow{4}{*}{OpenCLIP~\cite{openclip}} & ViT-B & \multirow{2}{*}{77.24} & $\mathcal{L}_{f_t}$ & 56.37 & 66.84 \\
& (57.26M, 11.27G) & & \cellcolor{Gray}$\mathcal{L}_{f_t} + \mathcal{L}_{\tilde{f_t}}$ & \cellcolor{Gray}\textbf{62.76 (+6.39)} & \cellcolor{Gray}\textbf{69.56 (+2.72)} \\
\cmidrule{2-6}
& ViT-L & \multirow{2}{*}{85.55} & $\mathcal{L}_{f_t}$ & 57.66 & 68.06 \\
& (202.05M, 51.89G) & & \cellcolor{Gray}$\mathcal{L}_{f_t} + \mathcal{L}_{\tilde{f_t}}$ & \cellcolor{Gray}\textbf{68.02 (+10.36)} & \cellcolor{Gray}\textbf{72.50 (+4.44)} \\

\bottomrule
\end{tabular}}}}
\end{center}
\vspace{-0.0cm}
\caption{Averaged image classification accuracy on art, clipart, and product using the student model pretrained on the images of real world as the source domain in OfficeHome.
The first and second numbers in the bracket of each model indicate the number of parameters and Multiply-Accumulate Operations (MACs), respectively.
* indicates that we performed linear probing using only labeled samples (\ie, images of source domain) for the teacher.}
\vspace{-0.3cm}
\label{tab:supple_uda_diverse} 
\end{table*}

\begin{table*}[t]
\centering
\begin{center}
{\resizebox{1.0\textwidth}{!}{
{
\begin{tabular}{c|c|cccccccccccc|c} 
\toprule
Metric & $\theta^c_t$ Init. & A2CA & A2P & A2RW & CA2A & CA2P & CA2RW & P2A & P2CA & P2RW & RW2A & RW2CA & RW2P & Avg. \\ 
\drule
\multirow{2}{*}{Teacher Acc.} & Random                    & 63.96 & 80.92 & 85.17 & 72.23 & 81.12 & 80.26 & 68.65 & 66.87 & 85.31 & 77.75 & 68.82 & 89.3 & 76.70 \\
             & \cellcolor{Gray}$\theta^c_s$                               & \cellcolor{Gray}65.59 & \cellcolor{Gray}79.95 & \cellcolor{Gray}84.51 & \cellcolor{Gray}70.29 & \cellcolor{Gray}80.15 & \cellcolor{Gray}79.66 & \cellcolor{Gray}65.68 & \cellcolor{Gray}64.72 & \cellcolor{Gray}82.76\cellcolor{Gray} & \cellcolor{Gray}76.02 & \cellcolor{Gray}66.76 & \cellcolor{Gray}89.64 & \cellcolor{Gray}75.48 \\
\midrule

\multirow{2}{*}{Student Acc.$\uparrow$} & Random                    & 32.71 & 43.59 & 58.30 & 35.15 & 52.89 & 50.91 & 30.00 & 33.33 & 59.65 & 53.36 & 44.81 & 71.07 & 47.15 \\
             & \cellcolor{Gray}$\theta^c_s$                               & \cellcolor{Gray}44.77 & \cellcolor{Gray}61.03 & \cellcolor{Gray}66.79 & \cellcolor{Gray}46.77 & \cellcolor{Gray}63.35 & \cellcolor{Gray}59.97 & \cellcolor{Gray}38.77 & \cellcolor{Gray}40.99 & \cellcolor{Gray}66.90 & \cellcolor{Gray}59.66 & \cellcolor{Gray}51.48 & \cellcolor{Gray}77.16 & \cellcolor{Gray}\textbf{56.47} \\

\midrule

\multirow{2}{*}{CKA($f_s$, $f_t$)$\uparrow$}& Random               & 0.29 & 0.50 & 0.47 & 0.36 & 0.55 & 0.47 & 0.34 & 0.42 & 0.49 & 0.42 & 0.50 & 0.59 & 0.45 \\
             & \cellcolor{Gray}$\theta^c_s$                               & \cellcolor{Gray}0.53 & \cellcolor{Gray}0.60 & \cellcolor{Gray}0.56 & \cellcolor{Gray}0.43 & \cellcolor{Gray}0.63 & \cellcolor{Gray}0.53 & \cellcolor{Gray}0.41 & \cellcolor{Gray}0.50 & \cellcolor{Gray}0.57 & \cellcolor{Gray}0.49 & \cellcolor{Gray}0.55 & \cellcolor{Gray}0.69 & \cellcolor{Gray}\textbf{0.54} \\

\midrule

\multirow{2}{*}{CKA($f_s$, $\tilde{f_t}$)$\uparrow$} & Random       & 0.31 & 0.50 & 0.49 & 0.34 & 0.51 & 0.42 & 0.33 & 0.41 & 0.46 & 0.44 & 0.49 & 0.58 & 0.44 \\
             & \cellcolor{Gray}$\theta^c_s$                               & \cellcolor{Gray}0.65 & \cellcolor{Gray}0.72 & \cellcolor{Gray}0.69 & \cellcolor{Gray}0.47 & \cellcolor{Gray}0.65 & \cellcolor{Gray}0.57 & \cellcolor{Gray}0.48 & \cellcolor{Gray}0.57 & \cellcolor{Gray}0.64 & \cellcolor{Gray}0.61 & \cellcolor{Gray}0.65 & \cellcolor{Gray}0.76 & \cellcolor{Gray}\textbf{0.62} \\

\bottomrule
\end{tabular}}}}
\end{center}
\caption{Comparisons on the initialization of the head classifier of teacher models during the feature customization stage. We report the results of each domain of Table~6 in the main paper. CKA refers to the centered kernel alignment~\cite{cka}. }
\label{tab:supple_uda_random} 
\end{table*}

\begin{table*}[t]
\centering
\begin{center}
{\resizebox{0.6\textwidth}{!}{
{
\begin{tabular}{ccc|ccc|c} 
\toprule
$\mathcal{L}_{L}$, $\mathcal{L}_{U}$ & \hspace{0.05cm} $\mathcal{L}_{f_t}$ & \hspace{0.02cm} $\mathcal{L}_{\tilde{f_t}}$ \hspace{0.025cm} & RW2A & RW2CA & RW2P & Avg. \\ 
\drule
\checkmark & \textendash & \textendash & 52.62 & 43.96 & 71.91 & 56.16 \\
\checkmark & \checkmark & \textendash  & 52.49 & 44.93 & 71.53 & 56.32 \\
\checkmark & \textendash & \checkmark  & 59.99 & 49.67 & 76.75 & 62.14 \\
\cellcolor{Gray}\checkmark & \cellcolor{Gray}\checkmark & \cellcolor{Gray}\checkmark & \cellcolor{Gray}59.66 & \cellcolor{Gray}51.50 & \cellcolor{Gray}77.13 & \cellcolor{Gray}\textbf{62.76} \\
\bottomrule
\end{tabular}

}}}
\end{center}
\caption{Ablation study on our loss functions. Given that the real world images are used for the source domain, we report the results of three target domains for Table~7(a) of the main paper.}
\label{tab:supple_uda_ablation_loss} 
\end{table*}

\begin{table*}[t]
\centering
\begin{center}
{\resizebox{0.6\textwidth}{!}{
{
\begin{tabular}{c|ccc|c} 
\toprule
Alternating Epochs & RW2A & RW2CA & RW2P & Avg. \\ 
\drule
30:1 & 59.25 & 50.74 & 76.68 & 62.22 \\
10:1 & 59.13 & 50.90 & 76.57 & 62.20 \\
5:1  & 59.29 & 51.02 & 76.53 & 62.28 \\
\cellcolor{Gray}1:1  & \cellcolor{Gray}59.66 & \cellcolor{Gray}51.50 & \cellcolor{Gray}77.13 & \cellcolor{Gray}\textbf{62.76} \\
\bottomrule
\end{tabular}

}}}
\end{center}
\caption{Analysis on the frequency of the feature customization stage. Given that the real world images are used for the source domain, we report the results of three target domains for Table~7(b) of the main paper.}
\vspace{-0.2cm}
\label{tab:supple_uda_ablation_epoch} 
\end{table*}

\end{document}